\documentclass[runningheads]{llncs}
\usepackage[T1]{fontenc}
\usepackage{graphicx}
\usepackage[table]{xcolor}
\usepackage{multirow}
\usepackage[pagebackref,breaklinks,colorlinks]{hyperref}

\usepackage{amsmath,amsfonts}

\begin{document}
\title{How to Squeeze An Explanation Out of Your Model\thanks{Accepted on xAI4Biometrics at ECCV 2024}}
%
%
\author{Tiago Roxo \and
Joana C. Costa \and
Pedro R. M. Inácio \and
Hugo Proença \\
}

%
\authorrunning{T. Roxo \textit{et al.}}
%
\institute{Instituto de Telecomunicações, Portugal and Universidade da Beira Interior, Portugal}

%
\maketitle              
%

\begin{abstract} 

Deep learning models are widely used nowadays for their reliability in performing various tasks. However, they do not typically provide the reasoning behind their decision, which is a significant drawback, particularly for more sensitive areas such as biometrics, security and healthcare. The most commonly used approaches to provide interpretability create visual attention heatmaps of regions of interest on an image based on models gradient backpropagation. Although this is a viable approach, current methods are targeted toward image settings and default/standard deep learning models, meaning that they require significant adaptations to work on video/multi-modal settings and custom architectures. This paper proposes an approach for interpretability that is model-agnostic, based on a novel use of the Squeeze and Excitation (SE) block that creates visual attention heatmaps. By including an SE block prior to the classification layer of any model, we are able to retrieve the most influential features via SE vector manipulation, one of the key components of the SE block. Our results show that this new SE-based interpretability can be applied to various models in image and video/multi-modal settings, namely biometrics of facial features with CelebA and behavioral biometrics using Active Speaker Detection datasets. Furthermore, our proposal does not compromise model performance toward the original task, and has competitive results with current interpretability approaches in state-of-the-art object datasets, highlighting its robustness to perform in varying data aside from the biometric context.


\end{abstract}


\section{Introduction}
\label{sec:intro}

Application scenarios for deep learning are diverse and steadily increasing everyday,
ranging from robotic~\cite{kleeberger2020survey}, person re-identification~\cite{ning2023occluded}, security~\cite{10510296,9786831}, anomaly detection~\cite{roxo2023exploring}, biometrics~\cite{cascone2023visual}, and even healthcare~\cite{guo2019deep}. One of the key reasons behind this widespread relates to the ability of deep learning to perform various tasks, given the appropriate training data. The main drawback of using these approaches is mainly linked to their black-box settings, \textit{i.e.} the lack of interpretability/explanations behind model decisions, which can be critical in more sensitive areas such as healthcare and security-related tasks. In particular, for biometric applications, explainability related to facial attributes is crucial for face verification systems, while assessment of behavioral biometrics, such as speaking mannerisms and non-verbal cues (body language), is also highly influential.

There has been an increased interest in interpretability/explainable\break approaches for deep learning in recent years. The most predominant technique relates to visual attention maps via backpropagation of model gradients~\cite{selvaraju2017grad} where the areas of more ``interest'' for the model are highlighted, providing an explicit idea of where the model is mostly sourcing outputs.
Other approaches not linked to visual interpretability are less common, although there are various works using game theoretic approaches~\cite{NIPS2017_7062} to explain the output of machine learning models, which are useful to assess models limitations and the most influential features~\cite{9502910}. Although this topic has gained interest over the years, most works mainly focus on gradient-based approaches in image settings and for standard deep learning models. This means that other settings (such as video/multi-modal) and custom models (variations of available implementations) are less explored, and currently available interpretability approaches do not work in such conditions or require heavy adaptations. 

\begin{figure}[!t]
\centering
\includegraphics[width=0.95\textwidth]{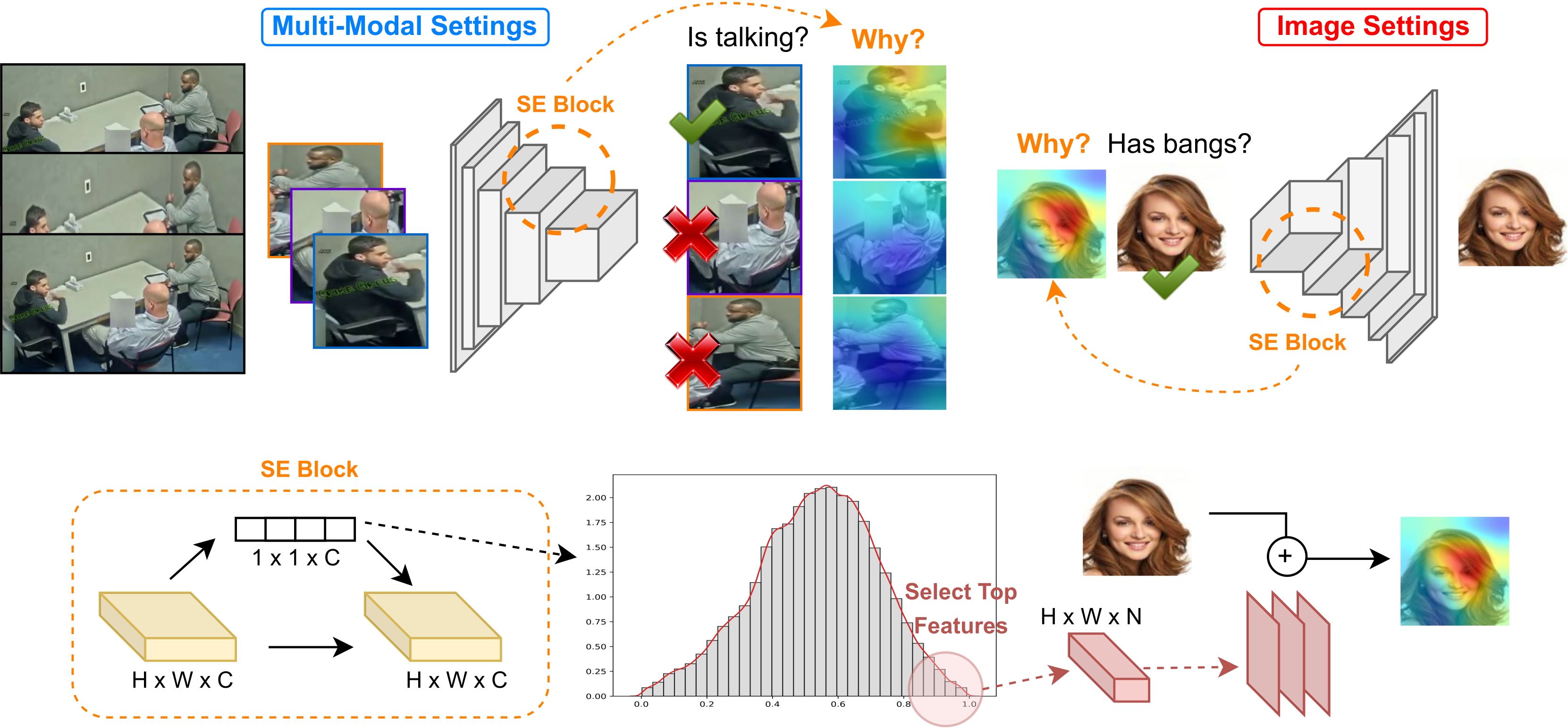}
\caption{
Overview of the inclusion of SE blocks for model-agnostic interpretability. Two different settings are depicted (multi-modal, on the left, and image, on the right) where SE block is similarly included in feature extractors to output attention heatmaps of models predictions. Heatmaps are created using channel features of the respective top 10\% SE vector values, via channel interpolation and combination with the original image. Audio encoding of the multi-modal setting (ASD) is not displayed for simplicity.
}
\label{fig:main_image}
\end{figure}

Given these limitations, we propose a novel use of the Squeeze-and-Excitation (SE) block~\cite{hu2018squeeze} to create a model-agnostic interpretability approach (Figure~\ref{fig:main_image}) that works regardless of the model, dataset and setting (image/video) used. SE block was initially proposed to account for the interdependencies between different feature channels to improve model performance. We adapted it to assess the most influential features to create visual heatmaps, similar to current interpretability approaches, via SE vector manipulation. Our proposal simply requires the inclusion of an SE block prior to the classification layer of deep learning models to be able to assess models attention. 
We explore the viability of SE-interpretability in state-of-the-art object datasets and in the biometric context, namely facial attributes and speaking behavioral features using CelebA~\cite{liu2015faceattributes} and Active Speaker Detection (ASD) data, respectively. The results show that the inclusion of SE blocks is a viable approach for visual interpretability, competitive with existing state-of-the-art approaches, and can be applied to different standard and custom models, using various datasets and settings, without degrading their performance on the original task.

The remaining of the paper is structured as follows. In Section 2, we
discuss the related work. In Section 3, we describe SE-related concepts and present our novel use of SE for interpretability. In Section 4 we report the experiments and discuss the results obtained. Finally, Section 5 concludes the paper.

\section{Related Work}
\label{sec:related-work}

\textbf{Model Interpretability.} For visual interpretability, we can group methods into two main categories: (i) \textit{gradient-based}~\cite{shrikumar2016not,srinivas2019full,smilkov2017smoothgrad,sundararajan2017axiomatic}, where gradients of each layer are computed through backpropagation; and (ii) \textit{attribution propagation}~\cite{gu2019understanding,iwana2019explaining,lundberg2017unified,shrikumar2017learning}, where there is a recursive decomposition of the contribution of the layers all the way to the input of the model. Other examples of visual interpretable approaches are \textit{Saliency-based methods}~\cite{zhou2018interpreting,dabkowski2017real}, which multiply the original image by the obtained saliency map to draw the focused region, \textit{Excitation Backprop}~\cite{zhang2018top}, which calculate a marginal winning probability by summing the values across channels, and \textit{Perturbation methods}~\cite{fong2019understanding,fong2017interpretable} that perturb the input and observe the changes made to the outputs. The increasing interest in Vision Transformers led to the \textit{Transformer-based interpretability}, inspired by standard techniques, such as gradient-based~\cite{chen2022lctr,gao2021ts,gupta2022vitol,qiang2022attcat}, attribution~\cite{chefer2021transformer,yuan2021explaining}, and redundancy reduction~\cite{pan2021ia}. Although most works explore visual interpretability in object classification datasets, its use in biometric-related topics such as face~\cite{yin2019towards,winter2022demystifying}, body~\cite{fu2021learning}, and Pedestrian Attribute Recognition (PAR)~\cite{doshi2023towards,9730882} data is not unprecedented.

\textbf{Video and Multi-Modal Interpretability.} The inclusion of GradCAM~\cite{selvaraju2017grad} and its variants is not commonly explored in video or multi-modal settings, aside from a few examples~\cite{stergiou2019saliency,hiley2020explaining}. Using GradCAM in video settings requires the video to be converted into a sequence of frames, ignoring the spatiotemporal correlations~\cite{roy2023explainable}. Furthermore, using the standard interpretability approaches in custom architectures, such as ASD context (multi-modal settings), is not extensively explored or smoothly done using the available implementations. Other approaches~\cite{bargal2018excitation,li2021towards} involve using a separate model/process to explain the decision of the original model. Our SE-based visual interpretation does not require additional computational cost or external methods to create attention heatmaps, is applicable for image and video/multi-modal settings, and can be included in standard and custom architectures.

\section{Squeeze-and-Excitation for Interpretability}
\label{sec:proposed_model}

\subsection{SE Definition and Concepts}
\label{sec:interpretability}

\begin{figure}[!t]
\centering
\includegraphics[width=0.99\textwidth]{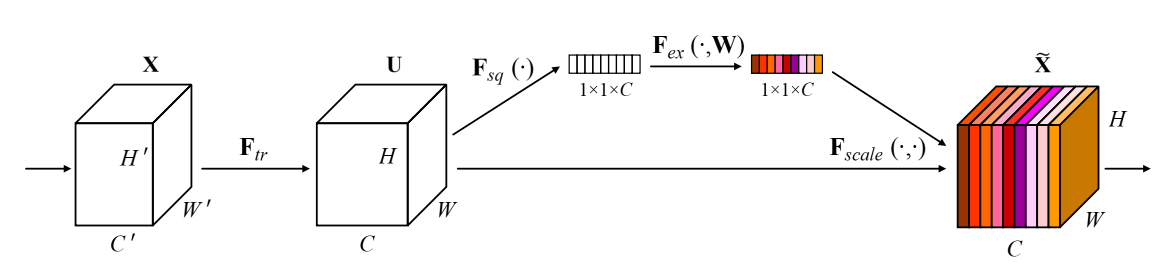}
\caption{
Overview of a Squeeze-and-Excitation block, where input \textit{U} is squeezed, for each channel ($u_c$, in (1)), via average pooling through spatial dimensions \textit{$H\times W$} to output a vector of weight importance ($\mathbf{s}$, in (2)), used to highlight the most important feature via channel-wise multiplication of weight importance vector and respective channel of the inputted feature map. Retrieved from the original paper~\cite{hu2018squeeze}.
}
\label{fig:se_block_orig}
\end{figure}

The original objective of SE blocks~\cite{hu2018squeeze} was to take into account the interdependencies between different feature channels to improve models performance, via three key stages: 1) Squeeze; 2) Excitation; and 3) Scale and Combine. 
Figure~\ref{fig:se_block_orig} presents the way SE blocks operate.  

\textbf{1) Squeeze Phase:} The goal of this phase is for the SE block to get a global understanding of each channel by squeezing them into a single numeric value, via global average pooling over the spatial dimensions (height $H$ and width $W$) of each channel. Formally, we can define $z \in  \mathbb{R}^{C}$, which is generated by shrinking input $U \in \mathbb{R}^{H\times W\times C}$ through its spatial dimensions $H \times W$, such that the $c$-\textit{th} element of $z$ is calculated by:
\begin{equation}
z_c = \frac{1}{H \times W} \sum_{H}^{i=1}\sum_{W}^{j=1} u_c(i,j).
\end{equation}

\textbf{2) Excitation Phase}: This phase aims to fully capture channel-wise dependencies of $z$, using  two fully-connected (FC), ReLU and Sigmoid layers, outputting vector $s$, with the same shape of $z$, representing the learned importance weights for each channel. Formally, $\mathbf{s}$ is defined by:
\begin{equation}
\mathbf{s} = \sigma (\mathbf{W}_2 \delta (\mathbf{W}_1 \mathbf{z})),
\end{equation}
where $\sigma$ refers to the sigmoid activation, $\delta$ is the ReLU fuction and $\mathbf{W}$ refer to the FC layers.

\textbf{3) Scale and Combine}: Finally, the weights of $s$ are incorporated into the input channels to highlight the most important features, via rescaling $U$ with 
$s$ using channel-wise multiplication between each channel of $s$ and the respective channel of the feature map $\mathbf{u}$, outputting $X \in \mathbb{R}^{H\times W\times C}$:
\begin{equation}
\mathbf{x}_c = s_c \mathbf{u}_c.
\end{equation}

\subsection{SE Visual Interpretability}

\begin{figure}[!tb]
\centering
\includegraphics[width=\textwidth]{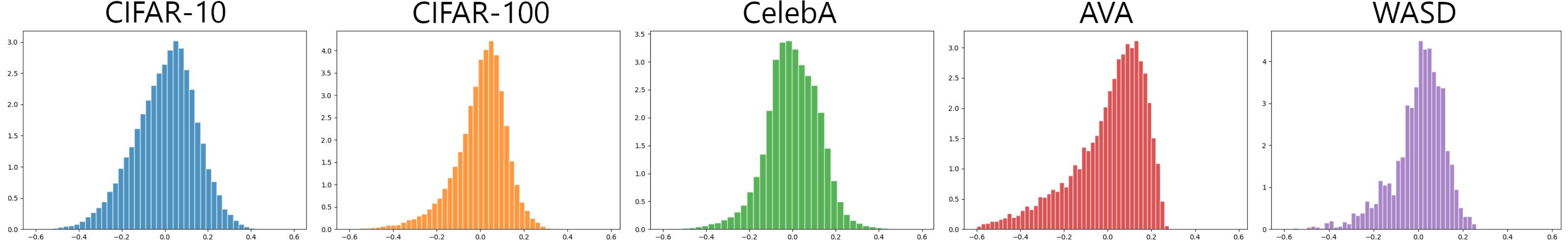}
\caption{
SE value distribution of all considered models for the datasets CIFAR-10, CIFAR-100, CelebA, AVA, and WASD (from left to right). SE values are normalized with mean to 0.
}
\label{fig:normal_distributions}
\end{figure}

We use SE to obtain the features (channels) perceived as highly influential (\textit{i.e.}, high value in SE vector, $\textbf{s}$ in (2)) for models to perform in different settings. However, the perception of \textit{high} is relative to the task considered, so we can not use a default/hard-coded value to define what is high or low importance. As such, given that the values of the SE vector in our experiments follow a Normal distribution (Figure~\ref{fig:normal_distributions}), we use its formula to obtain the top 10\% values of the SE vector (and, consequently, the top 10\% channels) for visual interpretability purposes. We obtain the normal distributions by grouping the SE values of all models for each dataset, by normalization with mean to 0. Finally, we conjugate the selected channels using bicubic interpolation on top of the original image to provide a heatmap of the most important regions for model decision.

\begin{figure}[!t]
\centering
\includegraphics[width=0.99\textwidth]{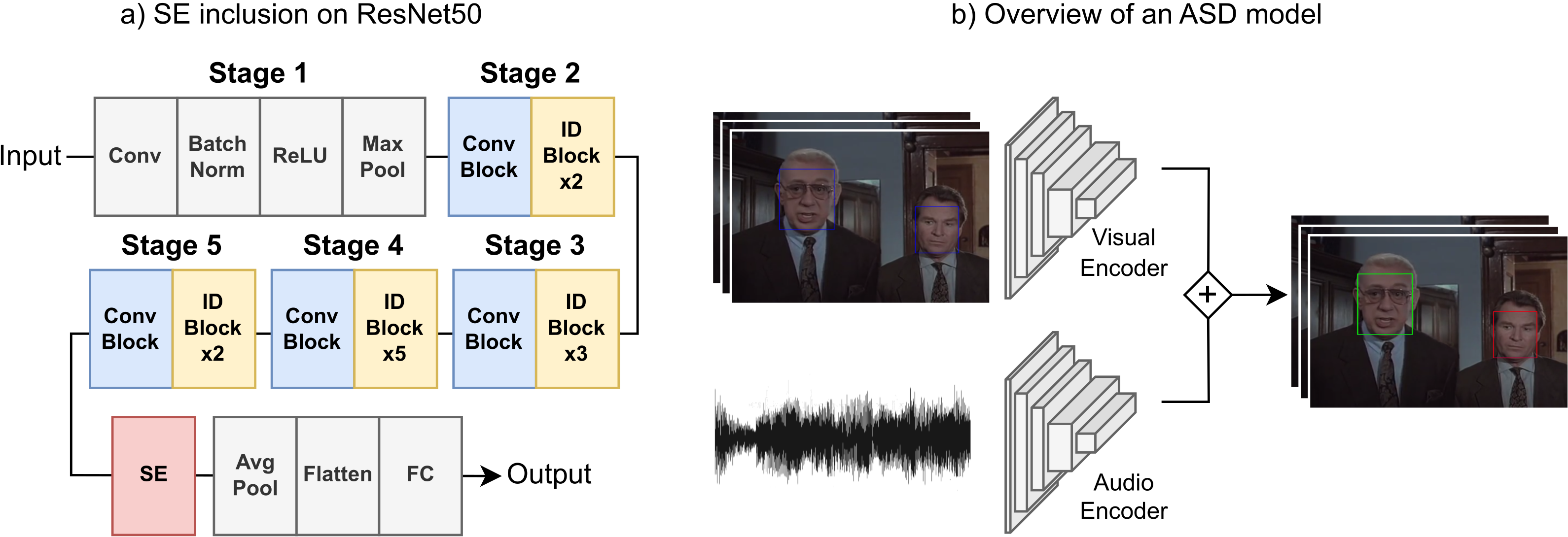}
\caption{
Inclusion of SE block in ResNet50 (left), and overview of how standard ASD models perform (right).
}
\label{fig:se_resnet_asd}
\end{figure}

\textbf{Motivation.} The novelty of our approach is the inclusion of a single SE block at the end of different models, in particular before average pooling (left side of Figure~\ref{fig:se_resnet_asd}) to provide visual interpretability similar to those obtained from GradCAM-like approaches. The key motivation is that inherently SE blocks already aim to assess the most influential channels to perform a given task, so we make use of the SE vector ($\textbf{s}$ in (2)) to select the features of the most influential channels ($\tilde{\textbf{X}}$, from Figure 2) to create model attention heatmaps. Given the simplicity of our proposal, SE blocks can provide visual interpretability to any model (standard or custom) in various settings (image or video-based).

\textbf{SE Interpretability Application.} SE interpretability is applicable to\break image-based data, as most interpretability-based approaches are able to perform. Additionally, it is also applicable in video/multi-modal settings, as we explore in our experiments using ASD as a proof-of-concept to assess behavioral biometrics. ASD models are commonly custom made (not standard ResNets) and use facial and audio data to predict who is talking, given a video input, which makes this area a prime candidate to assess SE interpretability robustness (right side of Figure~\ref{fig:se_resnet_asd}).

\subsection{Implementation Details}

Image models are trained using an SGD optimizer with a learning rate of 0.001, a momentum of 0.9, and a weight decay of 0.0005, during 50 epochs starting from pretrained weights on ImageNet. All inputs are resized to 128$\times$128, with the exception of Inceptionv3, which uses 299$\times$299 inputs due to its architecture constraints. Multi-modal models are trained with an Adam optimizer, with an initial learning rate of 10$^{-4}$, decreasing 5\% for each epoch. All visual data is reshaped into 112$\times$112, audio data is represented by 13-dimensional MFCC, and both visual and audio features have an encoding dimension of 128. For visual augmentation, we perform random flip, rotate and crop. For audio augmentation, we use negative audio sampling~\cite{tao2021someone}.

\section{Experiments}
\label{sec:experiments}

\subsection{Datasets, Methods, Models, and Evaluation Metrics}
\label{dataset}
 
\textbf{Datasets.} For image-based settings, we consider 3 datasets: CIFAR-10, CIFAR-100, and CelebA. CIFAR-10 dataset~\cite{krizhevsky2009learning} is a subset of the Tiny Images dataset~\cite{torralba2008tiny} and consists of 60 000 32$\times$32 color images, labeled into 10 mutually exclusive classes, each containing 6 000 images, with 5 000 training and 1 000 testing images per class. CIFAR-100 dataset~\cite{krizhevsky2009learning} contains the same data as CIFAR-10, except it has 100 classes with 600 images each, divided into 500 training images and 100 testing images per class. CelebA dataset~\cite{liu2015faceattributes} contains 202 599 face images with size 178$\times$218 from 10 177 celebrities, each annotated with 40 binary labels indicating facial attributes. In our experiments, we use CelebA for binary classification (only train models to predict one attribute).

Regarding multi-modal settings, we consider 2 ASD datasets:\break AVA-ActiveSpeaker~\cite{roth2020ava}, and WASD~\cite{10554644}. The AVA-ActiveSpeaker dataset~\cite{roth2020ava} is an audio-visual active speaker dataset from Hollywood movies, where typically only train and validation sets are used for experiments: 120 for training and 33 for validation, ranging from 1 to 10 seconds. WASD~\cite{10554644} compiles a set of videos from real interactions with varying accessibility of the two components for ASD: \textit{audio} and \textit{face}, containing 30 hours of labeled data, 164 videos, varying FPS, averaging 28-second duration, and balanced demographics.


\textbf{Interpretability Methods.} We compare the qualitative performance of our approach with GradCAM~\cite{selvaraju2017grad}, GradCAM++~\cite{chattopadhay2018grad}, EigenGradCAM~\cite{muhammad2020eigen}, and FullGradCAM~\cite{srinivas2019full} from publicly available implementations~\cite{jacobgilpytorchcam}.


\textbf{Models.} For image-based settings, the considered models are MobileNetv2, ResNet18, ResNet50 and InceptionV3, while for multi-modal conditions we select WASD Baseline~\cite{10554644} and Light-ASD~\cite{Liao_2023_CVPR}, given their end-to-end approach. Relative to Light-ASD, which only takes face and audio as inputs, WASD Baseline also considers body data as input.

\textbf{Evaluation Metrics.} For CIFAR-10, CIFAR-100, and CelebA we use accuracy, and for AVA-ActiveSpeaker and WASD, we use the official ActivityNet evaluation tool~\cite{roth2020ava} that computes mean Average Precision (mAP). For quantitative interpretability evaluation we use the Deletion and Insertion metric~\cite{Petsiuk2018rise}.

\begin{table}[!tb]
    \centering
    \small
    \renewcommand{\arraystretch}{1.05}
    \caption{SE inclusion performance (Acc) influence on CIFAR-10, CIFAR-100, and CelebA. Results on CelebA refer to gender attribute prediction. Par(M) refers to the number of parameters in millions.}
    \begin{tabular}{c|ccccc}\hline

        \makebox[8em]{\textbf{Model}} & \makebox[5em]{\textbf{Par(M)}} &
        \textbf{SE}
        & \makebox[6em]{\textbf{CIFAR-10}} & \makebox[6em]{\textbf{CIFAR-100}} & \makebox[6em]{\textbf{CelebA}}  \\
        \hline\hline
        
        \multirow{2}{*}{\makebox[5em]{MobileNetv2}} & \multirow{2}{*}{\makebox[5em]{2.13}} & $\times$ & 94.59 & 77.28 & 97.20 \\
        & & $\checkmark$ & 94.29 & 76.63 & 97.44 \\
        \hline

        \multirow{2}{*}{\makebox[5em]{ResNet18}} & \multirow{2}{*}{\makebox[5em]{10.66}} & $\times$ & 94.49 & 76.75 & 97.45 \\
        & & $\checkmark$ & 94.58 & 76.98 & 97.64 \\
        \hline

        \multirow{2}{*}{\makebox[5em]{ResNet50}} & \multirow{2}{*}{\makebox[5em]{22.44}} & $\times$ & 96.25 & 80.73 & 97.04 \\
        & & $\checkmark$ & 95.84 & 81.31 & 97.44 \\
        \hline

        \multirow{2}{*}{\makebox[5em]{Inceptionv3}} & \multirow{2}{*}{\makebox[5em]{23.97}} & $\times$ & 97.07 & 84.25 & 97.38 \\
        & & $\checkmark$ & 96.92 & 84.04 & 97.51 \\
        \hline
        \hline

    \end{tabular}
    \label{table:se-performance-object}
\end{table}

\begin{table}[!tb]
    \centering
    \small
    \renewcommand{\arraystretch}{1.05}
    \caption{SE inclusion performance (mAP) influence on AVA-ActiveSpeaker and WASD. AVA refers to AVA-ActiveSpeaker and Par(M) refers to the number of parameters in millions.}
    \begin{tabular}{c|cccccc}\hline

        \makebox[8em]{\textbf{Model}} & \makebox[8em]{\textbf{Visual Encoder}} & \makebox[3em]{\textbf{Par(M)}} &
        \makebox[3em]{\textbf{SE}}
        & \makebox[4em]{\textbf{AVA}} & \makebox[4em]{\textbf{WASD}} \\
        \hline\hline
        
        \multirow{2}{*}{\makebox[5em]{Light-ASD}~\cite{Liao_2023_CVPR}} & 
        \multirow{2}{*}{\makebox[5em]{Conv 2D-1D }} &
        \multirow{2}{*}{\makebox[5em]{1.0}} &$\times$ & 93.6 & 93.7 \\
        & & & $\checkmark$ & 93.7 & 93.9\\
        
        \hline
         
        \multirow{2}{*}{\makebox[5em]{Baseline}~\cite{10554644}} & 
        \multirow{2}{*}{\makebox[5em]{RN18 2D-3D}} &
        \multirow{2}{*}{\makebox[5em]{31.6}} &$\times$ & 92.3 & 94.0 \\
        & & & $\checkmark$ & 92.4 & 94.1 \\

        \hline
        \hline

    \end{tabular}
    \label{table:se-performance-multi-modal}
\end{table}

\subsection{SE Inclusion on Model Performance}

We start by assessing the effect of incorporating SE blocks into different architectures for image settings, in Table~\ref{table:se-performance-object}. The results show that using SE blocks can be applied in object and biometrics data, while not affecting the performance of models, regardless of their architectures and datasets, and without significantly increasing the number of model parameters. We also extend this evaluation to the multi-modal context, in particular ASD settings to assess behavioral biometrics, where audio and visual inputs are used for model predictions, in Table~\ref{table:se-performance-multi-modal}. The results confirm that SE blocks do not limit model performance, even in more challenging and variable conditions such as ASD data, showing its viability in behavioral biometrics related to speaking.

\begin{figure}[!tb]
\centering
\includegraphics[width=0.95\textwidth]{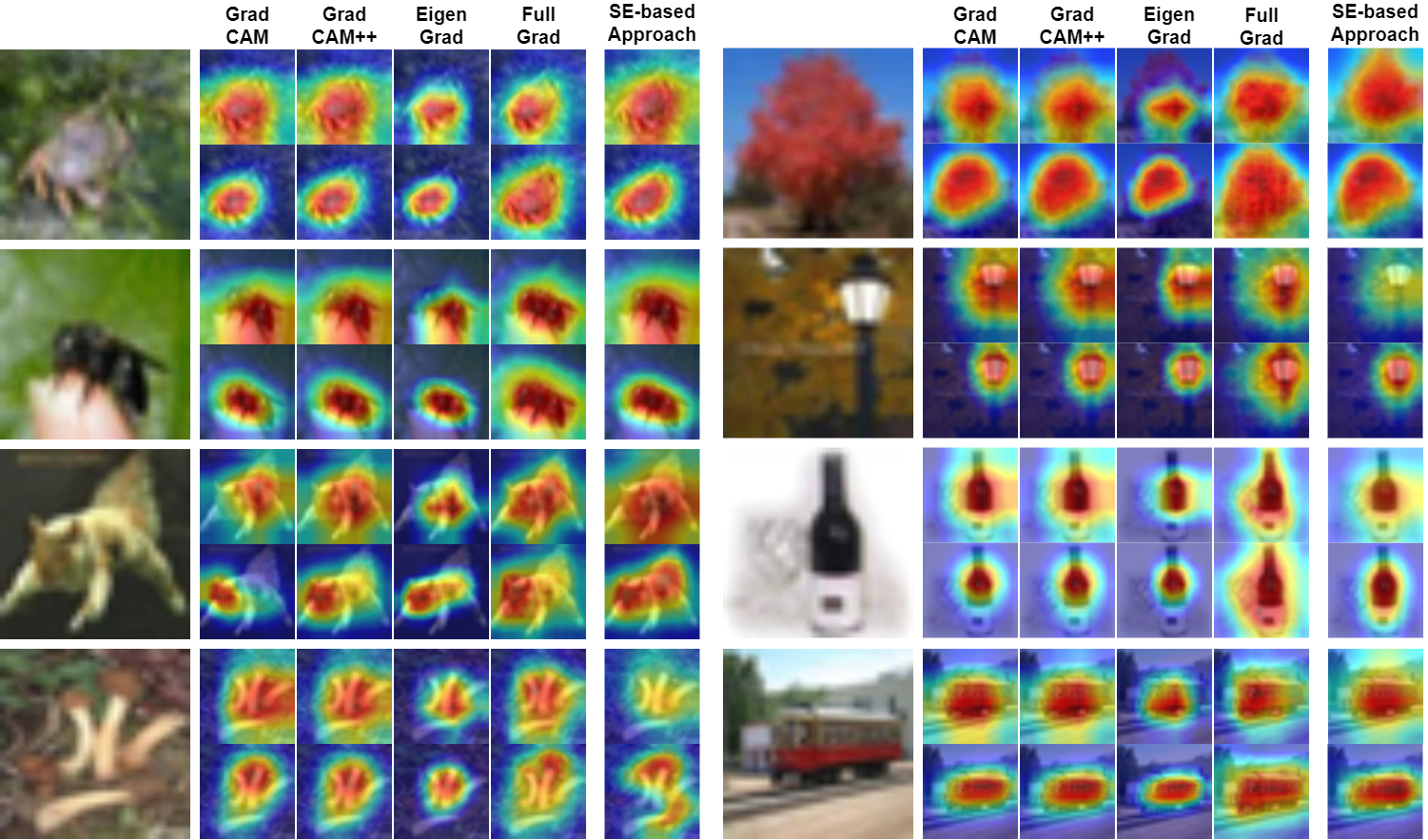}
\caption{
SE Interpretability (last column per original image) relative to other standard visual interpretability approaches: GradCAM, GradCAM++, EigenGrad, and FullGrad. For each original image the top row relates to interpretability of ResNet50 and the bottom to InceptionV3. 
}
\label{fig:cam_vs_se}
\end{figure}

\begin{table}[!tb]
    \centering
    \scriptsize
    \renewcommand{\arraystretch}{1.1}
    \caption{ResNet18 performance (Acc) on different attributes of CelebA.}
    \begin{tabular}{c|ccccccccc}\hline

        \multirow{2}{*}{\textbf{Attribute}} &
        \makebox[3em]{Arched} & 
        \multirow{2}{*}{\makebox[3em]{Bald}} & 
        \multirow{2}{*}{\makebox[3em]{Bangs}} &
        \makebox[3em]{Bushy} & 
        \makebox[3em]{Double} &
        \multirow{2}{*}{\makebox[4.5em]{Eyeglasses}} & 
        \multirow{2}{*}{\makebox[3em]{Beard}} &
        \multirow{2}{*}{\makebox[3em]{Earrings}} & 
        \multirow{2}{*}{\makebox[3em]{Hat}} \\

        & Eyebrows &  & & Eyebrows & Chin &&&\\
        
        \hline
        \hline
        \textbf{Acc} & 
        83.25 &
        98.65 & 
        95.75 &
        92.39 &
        96.32 & 
        99.63 &
        95.49 & 
        89.88 &
        98.99 \\

        \hline

    \end{tabular}
    \label{table:celebA-var}
\end{table}

\begin{figure}[!t]
\centering
\includegraphics[width=\textwidth]{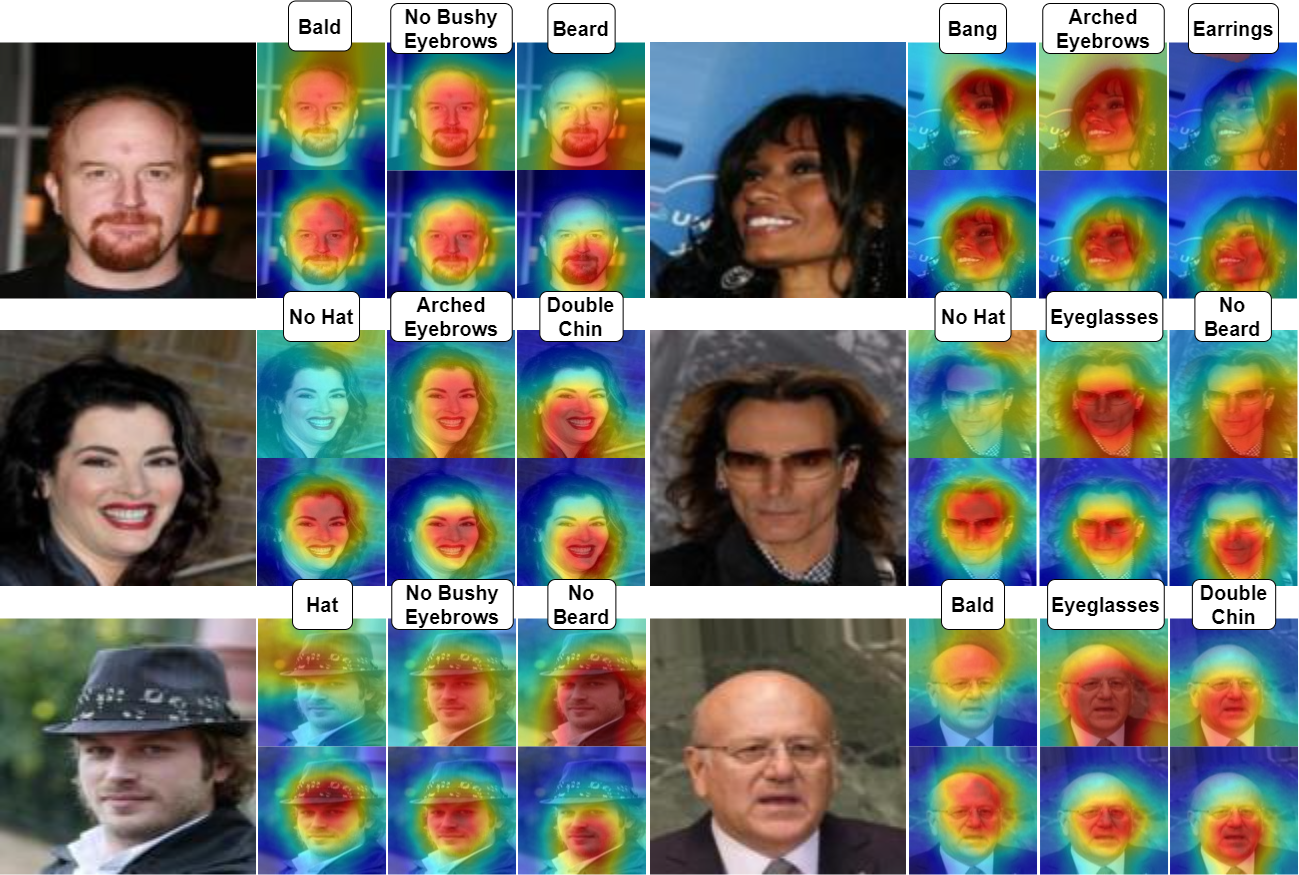}
\caption{
SE Interpretability of ResNet18 for different CelebA attributes. Each column after the original image refers to attributes more commonly located to specific parts of the face: top, middle, and bottom, respectively. For each original image the top row relates to interpretability of our approach (SE-based) and the bottom to FullGradCam.
}
\label{fig:celebA_var}
\end{figure}

\begin{figure}[!t]
\centering
\includegraphics[width=0.8\textwidth]{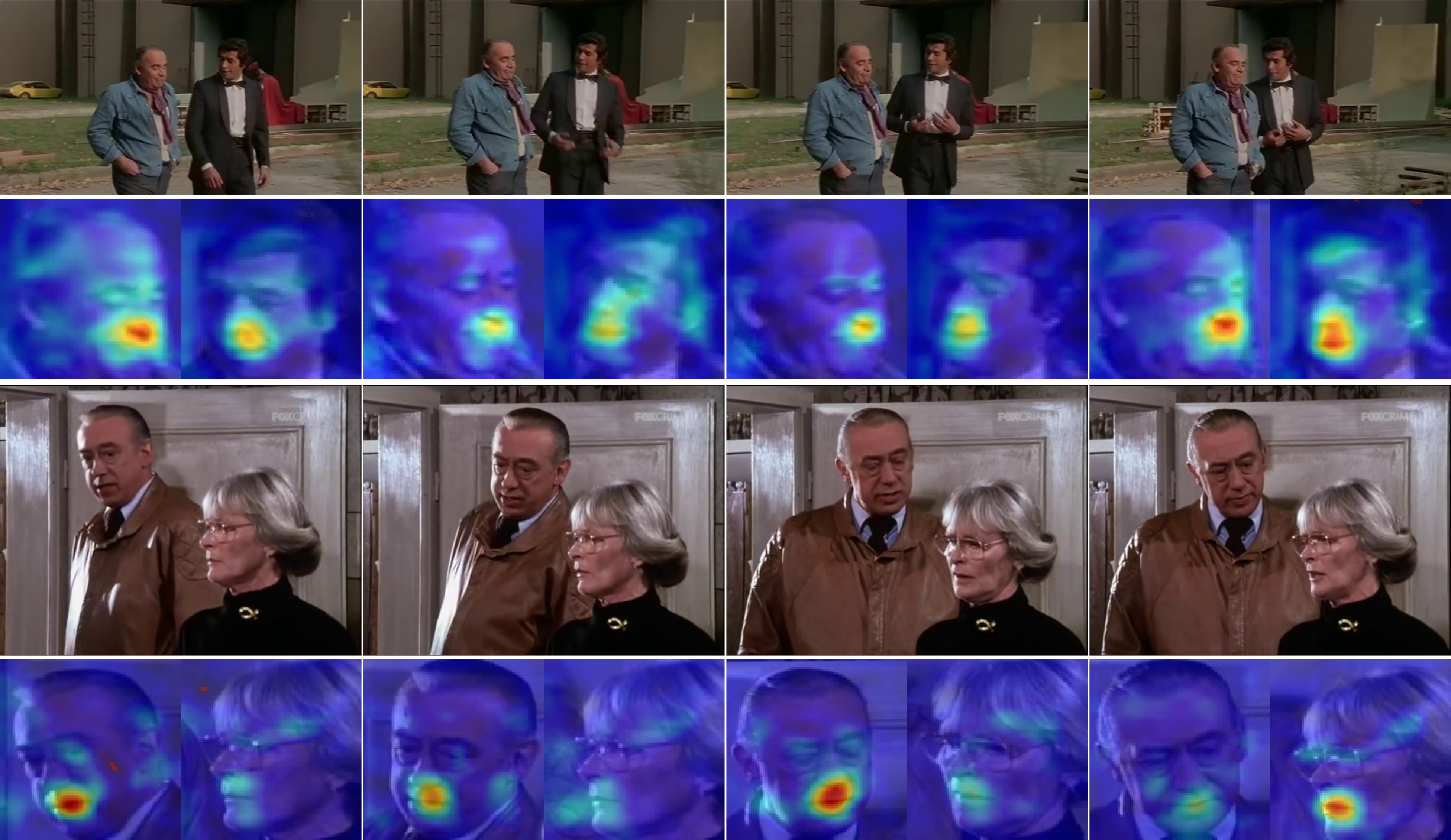}
\caption{
Attention of Light-ASD for facial cues, via SE blocks, in AVA-ActiveSpeaker data.
}
\label{fig:light_ava_attention}
\end{figure}

\subsection{Qualitative Performance of SE Interpretability}
\label{sec:qualitative}

Given that SE blocks do not influence model performance, regardless of architectures or training/testing settings, we assess the interpretability of models via SE blocks for CIFAR-100 in Figure~\ref{fig:cam_vs_se}. We use an object dataset to compare the viability of SE-interpretability relative to standard visual interpretability approaches (GradCAM variants), given the varying features available to classify in this data. The results show that our approach is able to output reliable and interpretable results for various settings: animals, plants, objects, with closed-up images, and varying positions of objects. Additionally, SE approach is applicable regardless of the architecture (ResNet \textit{vs.} Inception), although Inceptionv3 tends to have more focused attention, which is due to the input size requirements of this model (299x299 \textit{vs.} 128x128 of other models). We also assessed the influence of model attention in the biometric context, based on different attribute training using CelebA data, in Figure~\ref{fig:celebA_var} using our SE-based interpretability and FullGradCAM. The results show that our SE-based approach can be used to highlight the varying model attention based on the different attribute training: 1) hat presence, baldness or bangs relate to focus on top locations of the face; 2) eyeglasses and eyebrows lead to overall face attention; and 3) beard, accessories, and chin features relate to bottom locations of the face. Relative to our approach, FullGradCAM also highlights similar facial regions given different attributes, but with greater focus to the face (and less to the background). For reference, the performance of ResNet18 for each of the attributes assessed is shown in Table~\ref{table:celebA-var}.

\begin{figure}[!t]
\centering
\includegraphics[width=0.80\textwidth]{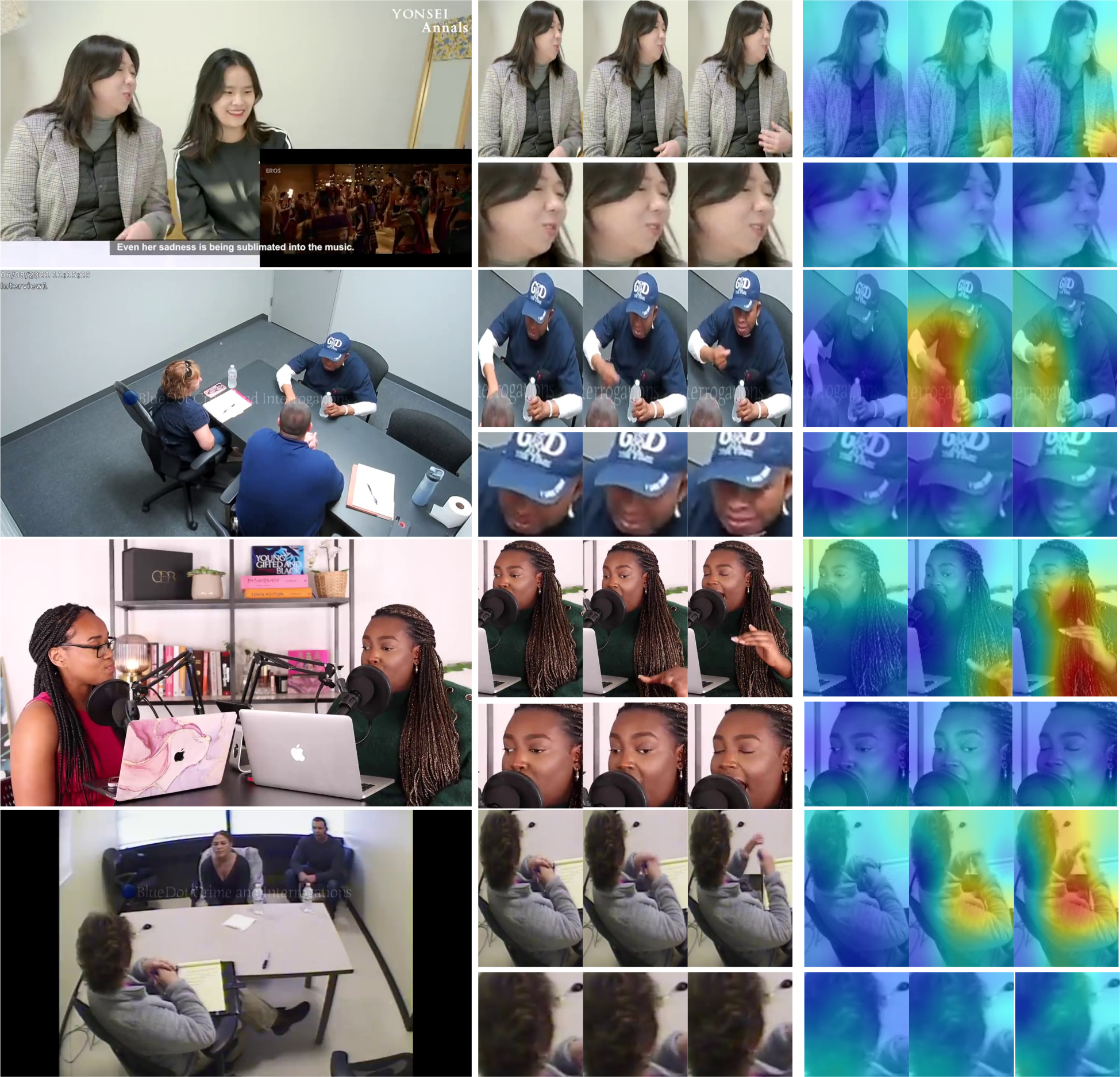}
\caption{
Body attention via SE blocks of WASD Baseline in challenging sets of WASD.
}
\label{fig:wasd_body_attention}
\end{figure}

\subsection{Multi-Modal Feature Importance via SE}

The relevance of SE blocks for interpretability also extends to other settings such as video/multi-modal, where we use ASD data in our experiments to assess the behavioral biometric context (facial attributes/cues analysis). Current ASD models take as input video facial data and associated audio to predict who are the active speakers in a scene. We start by exploring Light-ASD attention in different scenarios in Figure~\ref{fig:light_ava_attention}. The results show that the model focuses mainly on facial cues when people are talking, given that these are the most distinctive features, with greater attention to the person talking relative to other people in the scene, as seen in the second scenario. 

We also complement our experiments with WASD Baseline, which takes body data as an additional input, distinctive from other state-of-the-art ASD approaches, allowing for behavioral biometric assessment via body cue analysis (non-verbal speaking cues). We consider different scenarios of WASD to assess the relative importance of face and body attention heatmaps, in Figure~\ref{fig:wasd_body_attention}, where each scenario contains the initial setting (bigger image) and the next three frames for body and face movements (and their respective heatmaps). The results show greater attention to the body when the scenario contains challenging features, even if the face is available or partially accessible. In particular, with audio quality subpar (video playing, first row) or limited face availability (hat, second row), body movement importance increases for ASD. Furthermore, when limiting the access to facial cues (last two scenarios), the relative importance of the body further increases, translated by visual attention to hand movements (in both examples).

In these experiments, heatmaps for model attention tend to be consistent across consecutive frames and do not require big deviations (not talking and starts talking) to highlight model attention, although heatmap consistency across sequential frames may vary depending on the ASD context (\textit{i.e.} more cooperative ASD data leads to greater heatmap consistency). Given the results, our SE approach is a viable strategy to assess model attention in multi-modal settings, in particular for behavioral biometrics, regardless of the model architecture and dataset.

\begin{table}[!tb]
    \centering
    \small
    \renewcommand{\arraystretch}{1.05}
    \caption{Comparative evaluation in terms of Deletion (lower is better) and Insertion (higher is better) scores of a ResNet50 on CIFAR-10 and CIFAR-100 datasets.}
    \begin{tabular}{c|cccc}\hline

        \multirow{2}{*}{\makebox[10em]{\textbf{Approach}}} & \multicolumn{2}{c}{\textbf{CIFAR-10}} & \multicolumn{2}{c}{\textbf{CIFAR-100}} \\
         & \makebox[4em]{Deletion} & \makebox[4em]{Insertion} & \makebox[4em]{Deletion} & \makebox[4em]{Insertion} \\
        \hline\hline
        
        \makebox[5em]{GradCAM} & 0.31 & 0.93 & 0.17 & 0.77 \\

        \makebox[5em]{GradCAM$++$} & 0.30 & 0.93 & 0.15 & 0.78 \\

        \makebox[5em]{EigenGradCAM} & 0.29 & 0.93 & 0.14 & 0.78 \\

        \makebox[5em]{FullGradCAM} & 0.29 & 0.94 & 0.13 & 0.79\\

        \makebox[5em]{SE Interpretability} & 0.34 & 0.91 & 0.14 & 0.76 \\
        \hline
        \hline

    \end{tabular}
    \label{table:deletion_insertion_scores_rn50}
\end{table}

\subsection{Quantitative Performance of SE Interpretability}

\textbf{Deletion and Insertion.} To objectively compare our approach with standard interpretability methods, we use the Deletion and Insertion metric~\cite{Petsiuk2018rise} for ResNet50 in Table~\ref{table:deletion_insertion_scores_rn50}. \textit{Deletion} assesses the decrease in the probability of the predicted class as the most relevant pixels are removed (given the heatmaps), while \textit{Insertion} takes a complementary approach, \textit{i.e.} it measures the increase in probability as more and more pixels are introduced, with higher Area Under Curve indicative of a better explanation. The results show that our SE-based approach is competitive with standard interpretability approaches for CIFAR-10 and 100, with the advantage of being applicable to any model in various settings (image and video/multi-modal), making it a suitable option for model interpretability.

\begin{figure}[!tb]
\centering
\includegraphics[width=0.80\textwidth]{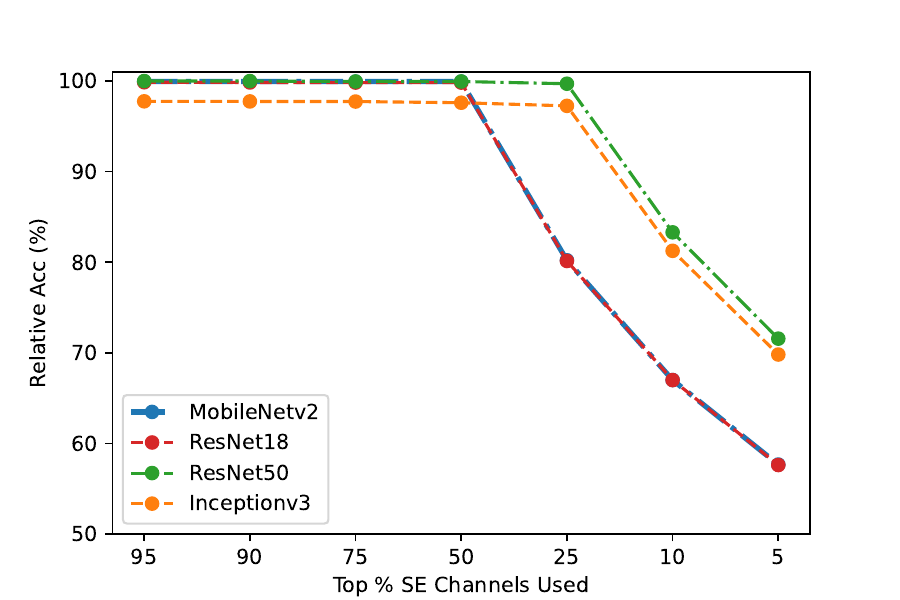}
\caption{
Accuracy with top \% of SE channels used in CIFAR-100, relative to using all available feature channels.
}
\label{fig:se-channel-var}
\end{figure}

\textbf{SE Channel Importance Variance.} We also explore the influence in performance when using feature channels corresponding to the top \% SE vector values, in Figure~\ref{fig:se-channel-var}. We select CIFAR-100 for this experiment given that it is the most challenging data in our experiments, based on the reported results in Table~\ref{table:se-performance-object} and their inherent challenges, as discussed in Section~\ref{sec:qualitative}. Although only the top percentage of SE channels are relevant to assess model attention, the decrease of feature channels used leads to performance decrease, with a greater emphasis using the top 25\% channels or less, for all considered models. Noticeably, ResNet50 and Inceptionv3 are more resilient to using fewer feature channels, given their performance with the top 25\% channels relative to MobileNetv2 and ResNet18, which is linked to model robustness (increased number of parameters).

\section{Conclusion}
\label{sec:conclusion}

In this paper we propose a novel application of Squeeze-and-Excitation (SE) blocks to assess model attention, with competitive qualitative results relative to common variants of GradCAM. The effectiveness of the proposed approach was proven in the biometrics context, in particular by assessing facial attributes and speaking features, via extensive testing with well known datasets both for images and videos, namely in the context of ASD. The interpretability provided by the SE blocks derives from its inherent intention to highlight the most discriminative features via SE vector, and can be applied to standard and custom architectures, in image or video/multi-modal settings, without compromising model performance. Future work includes exploring the effect of including SE blocks at different stages of a model to assess its attention to high (global) and low-level (local) features, and extend the experiments to assess whether the most salient zones are coherent with the predicted class using segmented face regions.

\begin{credits}
\subsubsection{\ackname} This work was supported by the Project UIDB/50008/2020, FCT Doctoral Grants 2020.09847.BD and 2021.04905.BD, and Project CENTRO-01-0145-FEDER-000019.

\subsubsection{\discintname}
The authors have no competing interests to declare that are relevant to the content of this article.
\end{credits}
%
%
%
%

\begin{thebibliography}{10}
\providecommand{\url}[1]{\texttt{#1}}
\providecommand{\urlprefix}{URL }
\providecommand{\doi}[1]{https://doi.org/#1}

\bibitem{bargal2018excitation}
Bargal, S.A., Zunino, A., Kim, D., Zhang, J., Murino, V., Sclaroff, S.: Excitation backprop for rnns. In: Proceedings of the IEEE conference on computer vision and pattern recognition. pp. 1440--1449 (2018)

\bibitem{cascone2023visual}
Cascone, L., Pero, C., Proen{\c{c}}a, H.: Visual and textual explainability for a biometric verification system based on piecewise facial attribute analysis. Image and Vision Computing  \textbf{132},  104645 (2023)

\bibitem{chattopadhay2018grad}
Chattopadhay, A., Sarkar, A., Howlader, P., Balasubramanian, V.N.: Grad-cam++: Generalized gradient-based visual explanations for deep convolutional networks. In: 2018 IEEE winter conference on applications of computer vision (WACV). pp. 839--847. IEEE (2018)

\bibitem{chefer2021transformer}
Chefer, H., Gur, S., Wolf, L.: Transformer interpretability beyond attention visualization. In: Proceedings of the IEEE/CVF Conference on CVPR. pp. 782--791 (2021)

\bibitem{chen2022lctr}
Chen, Z., Wang, C., Wang, Y., Jiang, G., Shen, Y., Tai, Y., Wang, C., Zhang, W., Cao, L.: Lctr: On awakening the local continuity of transformer for weakly supervised object localization. In: Proceedings of the AAAI Conference on Artificial Intelligence. vol.~36, pp. 410--418 (2022)

\bibitem{10510296}
Costa, J.C., Roxo, T., Proença, H., Inácio, P.R.M.: How deep learning sees the world: A survey on adversarial attacks \& defenses. IEEE Access  \textbf{12},  61113--61136 (2024). \doi{10.1109/ACCESS.2024.3395118}

\bibitem{9786831}
Costa, J.C., Roxo, T., Sequeiros, J.B.F., Proen{\c{c}}a, H., Inácio, P.R.M.: Predicting cvss metric via description interpretation. IEEE Access  \textbf{10},  59125--59134 (2022). \doi{10.1109/ACCESS.2022.3179692}

\bibitem{dabkowski2017real}
Dabkowski, P., Gal, Y.: Real time image saliency for black box classifiers. Advances in neural information processing systems  \textbf{30} (2017)

\bibitem{doshi2023towards}
Doshi, K., Yilmaz, Y.: Towards interpretable video anomaly detection. In: Proceedings of the IEEE/CVF WACV. pp. 2655--2664 (2023)

\bibitem{fong2019understanding}
Fong, R., Patrick, M., Vedaldi, A.: Understanding deep networks via extremal perturbations and smooth masks. In: Proceedings of the IEEE/CVF ICCV. pp. 2950--2958 (2019)

\bibitem{fong2017interpretable}
Fong, R.C., Vedaldi, A.: Interpretable explanations of black boxes by meaningful perturbation. In: Proceedings of the IEEE ICCV. pp. 3429--3437 (2017)

\bibitem{fu2021learning}
Fu, J., Gao, J., Xu, C.: Learning semantic-aware spatial-temporal attention for interpretable action recognition. IEEE Transactions on Circuits and Systems for Video Technology  \textbf{32}(8),  5213--5224 (2021)

\bibitem{gao2021ts}
Gao, W., Wan, F., Pan, X., Peng, Z., Tian, Q., Han, Z., Zhou, B., Ye, Q.: Ts-cam: Token semantic coupled attention map for weakly supervised object localization. In: Proceedings of the IEEE/CVF international conference on computer vision. pp. 2886--2895 (2021)

\bibitem{jacobgilpytorchcam}
Gildenblat, J., contributors: Pytorch library for cam methods. \url{https://github.com/jacobgil/pytorch-grad-cam} (2021)

\bibitem{gu2019understanding}
Gu, J., Yang, Y., Tresp, V.: Understanding individual decisions of cnns via contrastive backpropagation. In: Proceedings of the Computer Vision--ACCV 2018: 14th Asian Conference on Computer Vision, Perth, Australia, December 2--6, 2018, Revised Selected Papers, Part III 14. pp. 119--134. Springer (2019)

\bibitem{guo2019deep}
Guo, Z., Li, X., Huang, H., Guo, N., Li, Q.: Deep learning-based image segmentation on multimodal medical imaging. IEEE Transactions on Radiation and Plasma Medical Sciences  \textbf{3}(2),  162--169 (2019)

\bibitem{gupta2022vitol}
Gupta, S., Lakhotia, S., Rawat, A., Tallamraju, R.: Vitol: Vision transformer for weakly supervised object localization. In: Proceedings of the IEEE/CVF Conference on Computer Vision and Pattern Recognition. pp. 4101--4110 (2022)

\bibitem{hiley2020explaining}
Hiley, L., Preece, A., Hicks, Y., Chakraborty, S., Gurram, P., Tomsett, R.: Explaining motion relevance for activity recognition in video deep learning models. arXiv preprint arXiv:2003.14285  (2020)

\bibitem{hu2018squeeze}
Hu, J., Shen, L., Sun, G.: Squeeze-and-excitation networks. In: Proceedings of the IEEE conference on computer vision and pattern recognition. pp. 7132--7141 (2018)

\bibitem{iwana2019explaining}
Iwana, B.K., Kuroki, R., Uchida, S.: Explaining convolutional neural networks using softmax gradient layer-wise relevance propagation. In: Proceedings of the 2019 IEEE/CVF ICCVW. pp. 4176--4185. IEEE (2019)

\bibitem{kleeberger2020survey}
Kleeberger, K., Bormann, R., Kraus, W., Huber, M.F.: A survey on learning-based robotic grasping. Current Robotics Reports  \textbf{1},  239--249 (2020)

\bibitem{krizhevsky2009learning}
Krizhevsky, A., Hinton, G., et~al.: Learning multiple layers of features from tiny images. Master's thesis, University of Tront  (2009)

\bibitem{li2021towards}
Li, Z., Wang, W., Li, Z., Huang, Y., Sato, Y.: Towards visually explaining video understanding networks with perturbation. In: Proceedings of the IEEE/CVF Winter Conference on Applications of Computer Vision. pp. 1120--1129 (2021)

\bibitem{Liao_2023_CVPR}
Liao, J., Duan, H., Feng, K., Zhao, W., Yang, Y., Chen, L.: A light weight model for active speaker detection. In: Proceedings of the IEEE/CVF Conference on Computer Vision and Pattern Recognition (CVPR). pp. 22932--22941 (June 2023)

\bibitem{liu2015faceattributes}
Liu, Z., Luo, P., Wang, X., Tang, X.: Deep learning face attributes in the wild. In: Proceedings of International Conference on Computer Vision (ICCV) (December 2015)

\bibitem{NIPS2017_7062}
Lundberg, S.M., Lee, S.I.: A unified approach to interpreting model predictions. In: Guyon, I., Luxburg, U.V., Bengio, S., Wallach, H., Fergus, R., Vishwanathan, S., Garnett, R. (eds.) Advances in Neural Information Processing Systems 30, pp. 4765--4774. Curran Associates, Inc. (2017), \url{http://papers.nips.cc/paper/7062-a-unified-approach-to-interpreting-model-predictions.pdf}

\bibitem{lundberg2017unified}
Lundberg, S.M., Lee, S.I.: A unified approach to interpreting model predictions. Advances in neural information processing systems  \textbf{30} (2017)

\bibitem{muhammad2020eigen}
Muhammad, M.B., Yeasin, M.: Eigen-cam: Class activation map using principal components. In: 2020 international joint conference on neural networks (IJCNN). pp.~1--7. IEEE (2020)

\bibitem{ning2023occluded}
Ning, E., Wang, C., Zhang, H., Ning, X., Tiwari, P.: Occluded person re-identification with deep learning: a survey and perspectives. Expert Systems with Applications p. 122419 (2023)

\bibitem{pan2021ia}
Pan, B., Panda, R., Jiang, Y., Wang, Z., Feris, R., Oliva, A.: Ia-red2: Interpretability-aware redundancy reduction for vision transformers. Advances in Neural Information Processing Systems  \textbf{34},  24898--24911 (2021)

\bibitem{Petsiuk2018rise}
Petsiuk, V., Das, A., Saenko, K.: Rise: Randomized input sampling for explanation of black-box models. In: Proceedings of the British Machine Vision Conference (BMVC) (2018)

\bibitem{qiang2022attcat}
Qiang, Y., Pan, D., Li, C., Li, X., Jang, R., Zhu, D.: Attcat: Explaining transformers via attentive class activation tokens. Advances in neural information processing systems  \textbf{35},  5052--5064 (2022)

\bibitem{roth2020ava}
Roth, J., Chaudhuri, S., Klejch, O., Marvin, R., Gallagher, A., Kaver, L., Ramaswamy, S., Stopczynski, A., Schmid, C., Xi, Z., et~al.: Ava active speaker: An audio-visual dataset for active speaker detection. In: 2020 IEEE ICASSP. pp. 4492--4496. IEEE (2020)

\bibitem{10554644}
Roxo, T., Costa, J.C., Inácio, P.R.M., Proença, H.: Wasd: A wilder active speaker detection dataset. IEEE Transactions on Biometrics, Behavior, and Identity Science pp.~1--1 (2024). \doi{10.1109/TBIOM.2024.3412821}

\bibitem{roxo2023exploring}
Roxo, T., Costa, J.C., In{\'a}cio, P.R., Proen{\c{c}}a, H.: On exploring audio anomaly in speech. In: 2023 IEEE International Workshop on Information Forensics and Security (WIFS). pp.~1--6. IEEE (2023)

\bibitem{9502910}
Roxo, T., Proença, H.: Is gender “in-the-wild” inference really a solved problem? IEEE Transactions on Biometrics, Behavior, and Identity Science  \textbf{3}(4),  573--582 (2021). \doi{10.1109/TBIOM.2021.3100926}

\bibitem{9730882}
Roxo, T., Proença, H.: Yinyang-net: Complementing face and body information for wild gender recognition. IEEE Access  \textbf{10},  28122--28132 (2022). \doi{10.1109/ACCESS.2022.3157857}

\bibitem{roy2023explainable}
Roy, C., Nourani, M., Arya, S., Shanbhag, M., Rahman, T., Ragan, E.D., Ruozzi, N., Gogate, V.: Explainable activity recognition in videos using deep learning and tractable probabilistic models. ACM Transactions on Interactive Intelligent Systems  \textbf{13}(4),  1--32 (2023)

\bibitem{selvaraju2017grad}
Selvaraju, R.R., Cogswell, M., Das, A., Vedantam, R., Parikh, D., Batra, D.: Grad-cam: Visual explanations from deep networks via gradient-based localization. In: Proceedings of the IEEE ICCV. pp. 618--626 (2017)

\bibitem{shrikumar2017learning}
Shrikumar, A., Greenside, P., Kundaje, A.: Learning important features through propagating activation differences. In: Proceedings of the ICML. pp. 3145--3153. PMLR (2017)

\bibitem{shrikumar2016not}
Shrikumar, A., Greenside, P., Shcherbina, A., Kundaje, A.: Not just a black box: Learning important features through propagating activation differences. arXiv preprint arXiv:1605.01713  (2016)

\bibitem{smilkov2017smoothgrad}
Smilkov, D., Thorat, N., Kim, B., Vi{\'e}gas, F., Wattenberg, M.: Smoothgrad: removing noise by adding noise. arXiv preprint arXiv:1706.03825  (2017)

\bibitem{srinivas2019full}
Srinivas, S., Fleuret, F.: Full-gradient representation for neural network visualization. Advances in neural information processing systems  \textbf{32} (2019)

\bibitem{stergiou2019saliency}
Stergiou, A., Kapidis, G., Kalliatakis, G., Chrysoulas, C., Veltkamp, R., Poppe, R.: Saliency tubes: Visual explanations for spatio-temporal convolutions. In: 2019 IEEE international conference on image processing (ICIP). pp. 1830--1834. IEEE (2019)

\bibitem{sundararajan2017axiomatic}
Sundararajan, M., Taly, A., Yan, Q.: Axiomatic attribution for deep networks. In: Proceedings of the ICML. pp. 3319--3328. PMLR (2017)

\bibitem{tao2021someone}
Tao, R., Pan, Z., Das, R.K., Qian, X., Shou, M.Z., Li, H.: Is someone speaking? exploring long-term temporal features for audio-visual active speaker detection. In: Proceedings of the 29th ACM International Conference on Multimedia. pp. 3927--3935 (2021)

\bibitem{torralba2008tiny}
Torralba, A., Fergus, R., Freeman, W.T.: 80 million tiny images: A large data set for nonparametric object and scene recognition. IEEE Transactions on Pattern Analysis and Machine Intelligence  \textbf{30}(11),  1958--1970 (2008)

\bibitem{winter2022demystifying}
Winter, M., Bailer, W., Thallinger, G.: Demystifying face-recognition with locally interpretable boosted features (libf). In: 2022 10th EUVIP. pp.~1--6. IEEE (2022)

\bibitem{yin2019towards}
Yin, B., Tran, L., Li, H., Shen, X., Liu, X.: Towards interpretable face recognition. In: Proceedings of the IEEE/CVF ICCV. pp. 9348--9357 (2019)

\bibitem{yuan2021explaining}
Yuan, T., Li, X., Xiong, H., Cao, H., Dou, D.: Explaining information flow inside vision transformers using markov chain. In: eXplainable AI approaches for debugging and diagnosis. (2021)

\bibitem{zhang2018top}
Zhang, J., Bargal, S.A., Lin, Z., Brandt, J., Shen, X., Sclaroff, S.: Top-down neural attention by excitation backprop. International Journal of Computer Vision  \textbf{126}(10),  1084--1102 (2018)

\bibitem{zhou2018interpreting}
Zhou, B., Bau, D., Oliva, A., Torralba, A.: Interpreting deep visual representations via network dissection. IEEE TPAMI  \textbf{41}(9),  2131--2145 (2018)

\end{thebibliography}

\end{document}